\documentclass[11pt,a4paper]{article}
\usepackage[hyperref]{emnlp2018}
\usepackage{times}
\usepackage{latexsym}
\usepackage{url}
\usepackage{graphicx, amssymb, amsmath, amsfonts, color, ulem, xspace}
\usepackage[linesnumbered,ruled]{algorithm2e}
\usepackage[utf8]{inputenc}
\usepackage{flushend}
\usepackage{ulem}
\usepackage{siunitx}
\usepackage{booktabs}
\usepackage[russian,english]{babel}

\aclfinalcopy

\newcommand*{\bleu}[1]{\num[round-mode=places,round-precision=1]{#1}}

\title{Phrase-Based \& Neural Unsupervised Machine Translation}

\makeatletter
\renewcommand*{\@fnsymbol}[1]{\ensuremath{\ifcase#1\or \dagger\or \ddagger\or
    \mathsection\or \mathparagraph\or \|\or **\or \dagger\dagger
    \or \ddagger\ddagger \else\@ctrerr\fi}}
\makeatother

\author{
Guillaume Lample\thanks{\ Sorbonne Universit\'es, UPMC Univ Paris 06, CNRS, UMR 7606, LIP6, F-75005, Paris, France.} \\
Facebook AI Research \\
Sorbonne Universités \\
\texttt{glample@fb.com} \\
\And
Myle Ott \\
Facebook AI Research \\
\texttt{myleott@fb.com} \\
\And
Alexis Conneau \\
Facebook AI Research \\
Universit\'e Le Mans \\
\texttt{aconneau@fb.com} \\
\AND
Ludovic Denoyer$^{\dagger}$\\
Sorbonne Universités \\
\texttt{ludovic.denoyer@lip6.fr} \\
\And
Marc'Aurelio Ranzato \\
Facebook AI Research \\
\texttt{ranzato@fb.com} \\
}

\date{}

\begin{document}
\maketitle

\newcommand{\insertallresults}{
    \begin{table*}[h!]
    \small
    \begin{tabular}[b]{l|cc|cc|cc|cc}
    \toprule
    & en $\rightarrow$ fr  & fr$\rightarrow$ en & en$\rightarrow$ de & de$\rightarrow$ en & en$\rightarrow$ ro & ro$\rightarrow$ en & en$\rightarrow$ ru & ru$\rightarrow$ en \\
    \midrule
    \multicolumn{9}{l}{\it Unsupervised PBSMT} \\
    \midrule
    Unsupervised phrase table  &   -   & 17.50 &   -   & 15.63 &   -   & 14.10 &   -   &  8.08 \\
    Back-translation - Iter. 1 & 24.79 & 26.16 & 15.92 & 22.43 & 18.21 & 21.49 & 11.04 & 15.16 \\
    Back-translation - Iter. 2 & 27.32 & 26.80 & 17.65 & 22.85 & 20.61 & 22.52 & 12.87 & 16.42 \\
    Back-translation - Iter. 3 & 27.77 & 26.93 & 17.94 & 22.87 & 21.18 & 22.99 & 13.13 & 16.52 \\
    Back-translation - Iter. 4 & 27.84 & 27.20 & 17.77 & 22.68 & 21.33 & 23.01 & 13.37 & \textbf{16.62} \\
    Back-translation - Iter. 5 & \textbf{28.11} & 27.16 &   -   &   -   &   -   &   -   &   -   &   -   \\
    \midrule
    \multicolumn{9}{l}{\it Unsupervised NMT} \\
    \midrule
    LSTM                & 24.48 & 23.74 & 14.71 & 19.60 &   -   &   -   &   -   &   -   \\
    Transformer         & 25.14 & 24.18 & 17.16 & 21.00 & 21.18 & 19.44 &  7.98 &  9.09 \\
    \midrule
    \multicolumn{9}{l}{\it Phrase-based + Neural network} \\
    \midrule
    NMT + PBSMT         & 27.12 & 26.29 & 17.52 & 22.06 & 21.95 & 23.73 & 10.14 & 12.62 \\
    PBSMT + NMT         & 27.60 & \textbf{27.68} & \textbf{20.23} & \textbf{25.19} & \textbf{25.13} & \textbf{23.90} & \textbf{13.76} & \textbf{16.62} \\
    \bottomrule
    \end{tabular}
    \caption{\small \textbf{Fully unsupervised results.} We report the BLEU score for PBSMT, NMT, and their combinations on 8 directed language pairs. Results are obtained on \textit{newstest} 2014 for $en-fr$ and \textit{newstest} 2016 for every other pair.
    }
    \label{tab:allresults}
    \vspace{-0.25cm}
    \end{table*}
}

\newcommand{\insertmodelcomparisontable}{
    \begin{table}[tb]
    \small
    \begin{tabular}{l|cc|cc}
    \toprule
    Model & en-fr & fr-en & en-de & de-en \\
    \midrule
    \cite{unsupNMTartetxe}   & \bleu{15.13} & \bleu{15.56} &      -       &      -       \\
    \cite{unsupNMTlample}    & \bleu{15.04} & \bleu{14.31} & \bleu{09.64} & \bleu{13.33} \\
    \cite{unsupNMTyang}      & \bleu{16.97} & \bleu{15.58} & \bleu{10.86} & \bleu{14.62} \\
    \midrule
    NMT (LSTM)               & \bleu{24.48} & \bleu{23.74} & \bleu{14.71} & \bleu{19.60} \\
    NMT (Transformer)        & \bleu{25.14} & \bleu{24.18} & \bleu{17.16} & \bleu{21.00} \\
    PBSMT (Iter. 0)          & \bleu{16.17} & \bleu{17.50} & \bleu{10.98} & \bleu{15.63} \\
    PBSMT (Iter. n)          & \textbf{\bleu{28.11}} & \bleu{27.20} & \bleu{17.94} & \bleu{22.87} \\
    \midrule
    NMT + PBSMT              & \bleu{27.12} & \bleu{26.29} & \bleu{17.52} & \bleu{22.06} \\
    PBSMT + NMT              & \bleu{27.60} & \textbf{\bleu{27.68}} & \textbf{\bleu{20.23}} & \textbf{\bleu{25.19}} \\
    \bottomrule
    \end{tabular}
    \caption{\small \textbf{Comparison with previous approaches.} BLEU score for different models on the $en-fr$ and $en-de$ language pairs. Just using the unsupervised phrase table, and without back-translation (PBSMT (Iter. 0)), the PBSMT outperforms previous approaches. Combining PBSMT with NMT gives the best results.}
    \label{tab_comparison}
    \vspace{-0.5cm}
    \end{table}
}

\newcommand{\insertphrasetable}{
    \begin{table}[tb]
    \footnotesize
    \begin{tabular}{llcc}
    \toprule
    Source & Target & $P(\text{s}|\text{t})$ & $P(\text{t}|\text{s})$ \\
    \midrule
    \rule{0pt}{2.2ex}
            & happy & 0.931 & 0.986 \\
            & delighted & 0.458 & 0.003 \\
    heureux & grateful & 0.128 & 0.003 \\
            & thrilled & 0.392 & 0.002 \\
            & glad & 0.054 & 0.001 \\
    \midrule
    \rule{0pt}{2.2ex}
                & Britain & 0.242 & 0.720 \\
                & UK & 0.816 & 0.257 \\
    Royaume-Uni & U.K. & 0.697 & 0.011 \\
                & United Kingdom & 0.770 & 0.010 \\
                & British & 0.000 & 0.002 \\
    \midrule
    \rule{0pt}{2.2ex}
                       & European Union & 0.869 & 0.772 \\
                       & EU & 0.335 & 0.213 \\
    Union europ\'eenne & E.U. & 0.539 & 0.006 \\
                       & member states & 0.007 & 0.006 \\
                       & 27-nation bloc & 0.410 & 0.002 \\
    \bottomrule
    \end{tabular}
    
    \caption{\small \textbf{Unsupervised phrase table.} Example of candidate French to English phrase translations, along with their corresponding  conditional likelihoods.}
    \label{tab:phrase_table}
    \vspace{-0.6cm}
    \end{table}
}

\newcommand{\insertablationtable}{
    \begin{table}[!t]
    \centering
    \small
    \begin{tabular}{lcc}
    \toprule
    & en $\rightarrow$ fr & fr $\rightarrow$ en \\
    \midrule
    {\it Embedding Initialization} & & \\
    \midrule
    Concat + fastText (BPE) [default]  & \bleu{25.14} & \bleu{24.18} \\
    Concat + fastText (Words) & \bleu{20.97} & \bleu{20.89} \\
    fastText + Align (BPE)    & \bleu{22.04} & \bleu{21.27} \\
    fastText + Align (Words)  & \bleu{18.45} & \bleu{18.36} \\
    Random initialization     & \bleu{10.5} & \bleu{10.5} \\
    \midrule
    {\it Loss function} & & \\
    \midrule
    without $\mathcal{L}^{lm}$ of Eq.~\ref{eq:lmNMT}  &  \bleu{0.0} & \bleu{0.0} \\
    without $\mathcal{L}^{back}$ of Eq.~\ref{eq:btNMT}  &  \bleu{0.0} & \bleu{0.0} \\
    \midrule
    {\it Architecture} & & \\
    \midrule
    without sharing decoder & \bleu{24.6} & \bleu{23.7} \\
    LSTM instead of Transformer & \bleu{24.5} & \bleu{23.7} \\
    \bottomrule
    \end{tabular}
    
    \caption{\textbf{Ablation study of unsupervised NMT.} BLEU scores are computed over \textit{newstest} 2014. }
    \label{tab:abl}
    \vspace{-0.3cm}
    \end{table}
}

\newcommand{\inserttranslationsfren}{
    \begin{table*}[p]
    \resizebox{1\linewidth}{!}{
        \begin{tabular}{ll}
        \toprule
        \bf Source & \bf Je rêve constamment d'eux, peut-être pas toutes les nuits mais plusieurs fois par semaine c'est certain. \\
        \midrule
        NMT Epoch 1 & I constantly dream, but not all nights but by several times it is certain. \\
        NMT Epoch 3 & I continually dream them, perhaps not all but several times per week is certain. \\
        NMT Epoch 45 & I constantly dream of them, perhaps not all nights but several times a week it 's certain. \\
        \midrule
        PBSMT Iter. 0 & I dream of, but they constantly have all those nights but several times a week is too much. " \\
        PBSMT Iter. 1 & I had dreams constantly of them, probably not all nights but several times a week it is large. \\
        PBSMT Iter. 4 & I dream constantly of them, probably not all nights but several times a week it is certain. \\
        \midrule
        \bf Reference & \bf I constantly dream of them, perhaps not every night, but several times a week for sure. \\
        \midrule \\
        \midrule
        \bf Source & \bf La protéine que nous utilisons dans la glace réagit avec la langue à pH neutre. \\
        \midrule
        NMT Epoch 1 & The protein that we use in the ice with the language to pH. \\
        NMT Epoch 8 & The protein we use into the ice responds with language to pH neutral. \\
        NMT Epoch 45 & The protein we use in ice responds with the language from pH to neutral. \\
        \midrule
        PBSMT Iter. 0 & The protein that used in the ice responds with the language and pH neutral. \\
        PBSMT Iter. 1 & The protein that we use in the ice responds with the language to pH neutral. \\
        PBSMT Iter. 4 & The protein that we use in the ice reacts with the language to a neutral pH. \\
        \midrule
        \bf Reference & \bf The protein we are using in the ice cream reacts with your tongue at neutral pH. \\
        \midrule
        \\
        \midrule
        \bf Source & \bf Selon Google, les déguisements les plus recherchés sont les zombies, Batman, les pirates et les sorcières. \\
        \midrule
        NMT Epoch 1 & According to Google, there are more than zombies, Batman, and the pirates. \\
        NMT Epoch 8 & Google's most wanted outfits are the zombies, Batman, the pirates and the evil. \\
        NMT Epoch 45 & Google said the most wanted outfits are the zombies, Batman, the pirates and the witch. \\
        \midrule
        PBSMT Iter. 0 & According to Google, fancy dress and most wanted fugitives are the bad guys, Wolverine, the pirates and their minions. \\
        PBSMT Iter. 1 & According to Google, the outfits are the most wanted fugitives are zombies, Batman, pirates and witches. \\
        PBSMT Iter. 4 & According to Google, the outfits, the most wanted list are zombies, Batman, pirates and witches. \\
        \midrule
        \bf Reference & \bf According to Google, the highest searched costumes are zombies, Batman, pirates and witches. \\ 
        \bottomrule
        \end{tabular}
        \smallskip
    }
    \caption{\small \textbf{Unsupervised translations.} Examples of translations on the French-English pair of \textit{newstest} 2014 at different iterations of training. For PBSMT, we show translations at iterations 0, 1 and 4, where the model obtains BLEU scores of 15.4, 23.7 and 24.7 respectively. For NMT, we show examples of translations after epochs 1, 8 and 42, where the model obtains BLEU scores of 12.3, 17.5 and 24.2 respectively. Iteration 0 refers to the PBSMT model obtained using the unsupervised phrase table, and an epoch corresponds to training the NMT model on 500k monolingual sentences. At the end of training, both models generate very good translations.
    \label{tab:translations}}
    \end{table*}
}

\newcommand{\inserttranslationsruen}{
    \begin{table*}[p]
    \resizebox{1\linewidth}{!}{
        \begin{tabular}{ll}
        \toprule
        \multicolumn{2}{l}{\textbf{\textit{Russian $\rightarrow$ English}}}\\
        \midrule
        Source     & \begin{otherlanguage*}{russian}Изменения предусматривают сохранение льготы на проезд в общественном пассажирском транспорте.\end{otherlanguage*} \\
        Hypothesis & The changes involve keeping the benefits of parking in a public passenger transportation. \\
        Reference  & These changes make allowances for the preservation of discounted travel on public transportation. \\
        \midrule
        Source     & \begin{otherlanguage*}{russian}Шесть из 10 республиканцев говорят, что они согласны с Трампом по поводу иммиграции.\end{otherlanguage*} \\
        Hypothesis & Six in 10 Republicans say that they agree with Trump regarding immigration. \\
        Reference  & Six in 10 Republicans say they agree with Trump on immigration. \\
        \midrule
        Source     & \begin{otherlanguage*}{russian}Metcash пытается защитить свои магазины IGA от натиска Aldi в Южной Австралии и Западной Австралии .\end{otherlanguage*} \\
        Hypothesis & Metcash is trying to protect their shops IGA from the onslaught of Aldi in South Australia and Western Australia. \\
        Reference  & Metcash is trying to protect its IGA stores from an Aldi onslaught in South Australia and Western Australia. \\
        \midrule
        Source     & \begin{otherlanguage*}{russian}В них сегодня работают четыре сотни студентов из столичных колледжей и вузов.\end{otherlanguage*} \\
        Hypothesis & Others today employs four hundreds of students from elite colleges and universities. \\
        Reference  & Four hundred students from colleges and universities in the capital are working inside of it today. \\
        \bottomrule

        \end{tabular}
        \smallskip
    }
    \caption{\small \textbf{Unsupervised translations: Russian-English.} Examples of translations on the Russian-English pair of \textit{newstest} 2016 using the PBSMT model. \label{tab:translationsRu}}
    \end{table*}
}

\newcommand{\inserttranslationsdeen}{
    \begin{table*}[p]
    \resizebox{1\linewidth}{!}{
        \begin{tabular}{ll}
        \toprule
        \multicolumn{2}{l}{\textbf{\textit{German $\rightarrow$ English}}}\\
    
        \midrule
    
        Source    & Flüchtlinge brauchen Unterkünfte : Studie warnt vor Wohnungslücke \\
        PBSMT     & Refugees need accommodation : Study warns Wohnungslücke \\
        NMT       & Refugees need forestry study to warn housing gap \\
        PBSMT+NMT & Refugees need accommodation : Study warns of housing gap \\
        Reference & Refugees need accommodation : study warns of housing gap \\
    
        \midrule
    
        Source    & Konflikte : Mehrheit unterstützt die Anti-IS-Ausbildungsmission \\
        PBSMT     & Conflict : Majority supports Anti-IS-Ausbildungsmission \\
        NMT       & Tensions support majority anti-IS-recruiting mission \\
        PBSMT+NMT & Tensions : Majority supports the anti-IS-recruitment mission \\
        Reference & Conflicts : the majority support anti ISIS training mission \\
    
        \midrule
    
        Source    & Roboterautos : Regierung will Vorreiterrolle für Deutschland \\
        PBSMT     & Roboterautos : Government will step forward for Germany \\
        NMT       & Robotic cars will pre-reiterate government for Germany \\
        PBSMT+NMT & Robotic cars : government wants pre-orders for Germany \\
        Reference & Robot cars : Government wants Germany to take a leading role \\
    
        \midrule
    
        Source    & Pfund steigt durch beschleunigtes Lohnwachstum im Vereinigten Königreich \\
        PBSMT     & Pound rises as UK beschleunigtes Lohnwachstum . \\
        NMT       & £ rises through rapid wage growth in the U.S. \\
        PBSMT+NMT & Pound rises by accelerating wage growth in the United Kingdom \\
        Reference & Pound rises as UK wage growth accelerates \\
    
        \midrule
    
        Source    & 46 Prozent sagten , dass sie die Tür offen lassen , um zu anderen Kandidaten zu wechseln . \\
        PBSMT     & 52 per cent said that they left the door open to them to switch to other candidates . \\
        NMT       & 46 percent said that they would let the door open to switch to other candidates . \\
        PBSMT+NMT & 46 percent said that they left the door open to switch to other candidates . \\
        Reference & 46 percent said they are leaving the door open to switching candidates . \\
    
        \midrule
    
        Source    & Selbst wenn die Republikaner sich um einen anderen Kandidaten sammelten , schlägt Trump noch fast jeden . \\
        PBSMT     & Even if the Republicans a way to other candidates collected , beats Trump , yet almost everyone . \\
        NMT       & Even if Republicans are to donate to a different candidate , Trump takes to almost every place . \\
        PBSMT+NMT & Even if Republicans gather to nominate another candidate , Trump still beats nearly everyone . \\
        Reference & Even if Republicans rallied around another candidate , Trump still beats almost everyone . \\
    
        \midrule
    
        Source    & Ich glaube sicher , dass es nicht genügend Beweise gibt , um ein Todesurteil zu rechtfertigen . \\
        PBSMT     & I think for sure that there was not enough evidence to justify a death sentence . \\
        NMT       & I believe it 's not sure there 's enough evidence to justify a executions . \\
        PBSMT+NMT & I sure believe there is not enough evidence to justify a death sentence . \\
        Reference & I certainly believe there was not enough evidence to justify a death sentence . \\
    
        \midrule
    
        Source    & Auch wenn der Laden gut besucht ist , ist es nicht schwer , einen Parkplatz zu finden . \\
        PBSMT     & Even if the store visited it is , it is not hard to find a parking lot . \\
        NMT       & To be sure , the shop is well visited , but it 's not a troubled driveway . \\
        PBSMT+NMT & Even if the shop is well visited , it is not hard to find a parking lot . \\
        Reference & While the store can get busy , parking is usually not hard to find . \\
    
        \midrule
    
        Source    & Die Suite , in dem der kanadische Sänger wohnt , kostet am Tag genau so viel , wie ihre Mama Ewa im halben Jahr verdient . \\
        PBSMT     & The suite in which the Canadian singer grew up , costs on the day so much as their mum Vera in half year earned . \\
        NMT       & The Suite , in which the Canadian singer lived , costs day care exactly as much as her mom Ewa earned during the decade . \\
        PBSMT+NMT & The suite , in which the Canadian singer lived , costs a day precisely as much as her mom Ewa earned in half . \\
        Reference & The suite where the Canadian singer is staying costs as much for one night as her mother Ewa earns in six months . \\
    
        \midrule
    
        Source    & Der durchschnittliche BMI unter denen , die sich dieser Operation unterzogen , sank von 31 auf 24,5 bis Ende des fünften Jahres in dieser Studie . \\
        PBSMT     & The average BMI among those who undergoing this operation decreased by 22 to 325 by the end of fifth year in this study . \\
        NMT       & The average BMI among those who undergo surgery , sank from 31 to 24,500 by the end of the fifth year in that study . \\
        PBSMT+NMT & The average BMI among those who undergo this surgery fell from 31 to 24.5 by the end of the fifth year in this study . \\
        Reference & The average BMI among those who had surgery fell from 31 to 24.5 by the end of their fifth year in the study . \\
    
        \midrule
    
        Source    & Die 300 Plakate sind von Künstlern , die ihre Arbeit im Museum für grausame Designs in Banksys Dismaland ausgestellt haben . \\
        PBSMT     & The 250 posters are by artists and their work in the museum for gruesome designs in Dismaland Banksys have displayed . \\
        NMT       & The 300 posters are posters of artists who have their work Museum for real-life Designs in Banksys and Dismalausgestellt . \\
        PBSMT+NMT & The 300 posters are from artists who have displayed their work at the Museum of cruel design in Banksys Dismaland . \\
        Reference & The 300 posters are by artists who exhibited work at the Museum of Cruel Designs in Banksy 's Dismaland . \\
    
        \midrule
    
        Source    & Bis zum Ende des Tages gab es einen weiteren Tod : Lamm nahm sich das Leben , als die Polizei ihn einkesselte . \\
        PBSMT     & At the end of the day it 's a further death : Lamb took out life as police him einkesselte . \\
        NMT       & By the end of the day there was another death : Lamm emerged this year as police were trying to looseless him . \\
        PBSMT+NMT & By the end of the day , there 's another death : Lamm took out life as police arrived to him . \\
        Reference & By the end of the day , there would be one more death : Lamb took his own life as police closed in on him . \\
    
        \midrule
    
        Source    & Chaos folgte an der Grenze , als Hunderte von Migranten sich in einem Niemandsland ansammelten und serbische Beamte mit Empörung reagierten . \\
        PBSMT     & Chaos followed at the border , as hundreds of migrants in a frontier ansammelten and Serb officers reacted with outrage . \\
        NMT       & Chaos followed an avalanche as hundreds of thousands of immigrants fled to a mandsdom in answerage and Serbian officers responded with outrage . \\
        PBSMT+NMT & Chaos followed at the border as hundreds of immigrants gathered in a bush town and Serb officers reacted with outrage . \\
        Reference & Chaos ensued at the border , as hundreds of migrants piled up in a no man 's land , and Serbian officials reacted with outrage . \\
    
        \midrule
    
        Source    & " Zu unserer Reise gehörten viele dunkle Bahn- und Busfahrten , ebenso Hunger , Durst , Kälte und Angst " , schrieb sie . \\
        PBSMT     & " To our trip included many of the dark rail and bus rides , such as hunger , thirst , cold and fear , " she wrote . \\
        NMT       & " During our trip , many included dark bus and bus trips , especially hunger , Durst , and cold fear , " she wrote . \\
        PBSMT+NMT & " Our trip included many dark rail and bus journeys , as well as hunger , thirst , cold and fear , " she wrote . \\
        Reference & " Our journey involved many dark train and bus rides , as well as hunger , thirst , cold and fear , " she wrote . \\
    
        \bottomrule
    
        \end{tabular}
        \smallskip
    }
    \caption{\small \textbf{Unsupervised translations: German-English.} Examples of translations on the German-English pair of \textit{newstest} 2016 using the PBSMT, NMT, and PBSMT+NMT. \label{tab:translations_deen}}
    \end{table*}
}

\begin{abstract}
Machine translation systems achieve near human-level performance on some languages, yet
their effectiveness strongly relies on the availability of large amounts of parallel sentences, which hinders their applicability to the majority of language pairs. This work investigates how to learn to translate when having access to only large monolingual corpora in each language. We propose two model variants, a neural and a phrase-based model. Both versions leverage a careful initialization of the parameters, the denoising effect of language models and automatic generation of parallel data by iterative back-translation. These models are significantly better than methods from the literature, while being simpler and having fewer hyper-parameters.
On the widely used WMT'14 English-French and WMT'16 German-English benchmarks, our models respectively obtain 28.1 and 25.2 BLEU points without using a single parallel sentence, outperforming the state of the art by more than 11 BLEU points. On low-resource languages like English-Urdu and English-Romanian, our methods achieve even better results than semi-supervised and supervised approaches leveraging the paucity of available bitexts. Our code for NMT and PBSMT is publicly available.\footnote{\url{https://github.com/facebookresearch/UnsupervisedMT}}\end{abstract}

\begin{figure*}[t!]
\includegraphics[width=\linewidth]{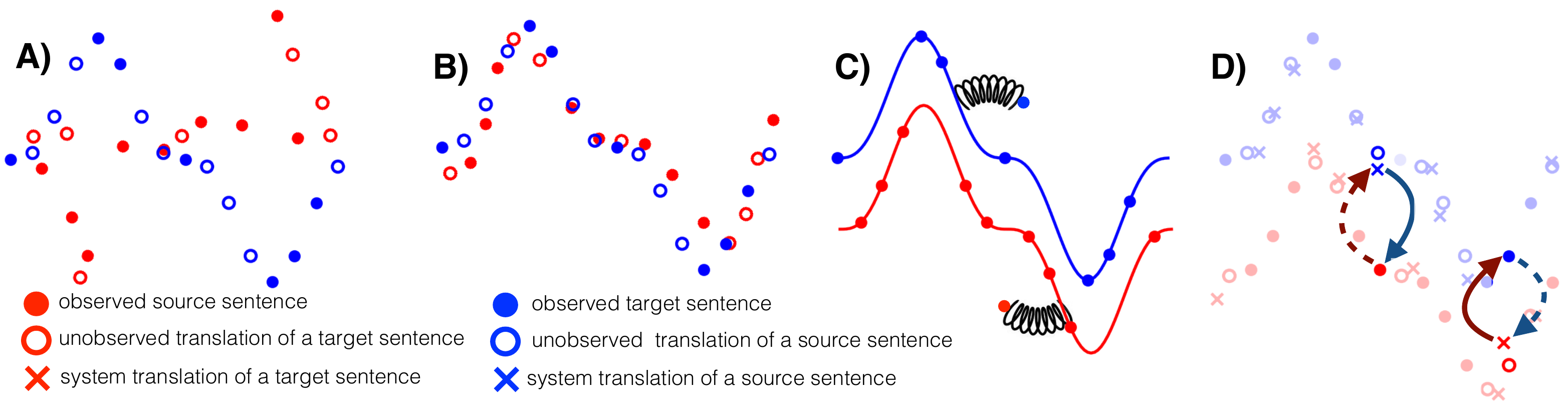}
\vspace{-0.8cm}
\caption{\small{Toy illustration of the three principles of unsupervised MT. \textbf{A)} There are two monolingual datasets. Markers correspond to sentences (see legend for details). \textbf{B)} First principle: \textbf{Initialization}. The two distributions are roughly aligned, e.g. by performing word-by-word translation with an inferred bilingual dictionary. \textbf{C)} Second principle: \textbf{Language modeling}. A language model is learned independently in each domain to infer the structure in the data (underlying continuous curve); it acts as a data-driven prior to denoise/correct sentences (illustrated by the spring pulling a sentence outside the manifold back in). \textbf{D)} Third principle: \textbf{Back-translation}. Starting from an observed source sentence (filled red circle) we use the current source~$\rightarrow$~target model to translate (dashed arrow), yielding a potentially incorrect translation (blue cross near the empty circle). Starting from this (back) translation, we use the target~$\rightarrow$~source model (continuous arrow) to reconstruct the sentence in the original language. The discrepancy between the reconstruction and the initial sentence provides error signal to train the target~$\rightarrow$~source model parameters. The same procedure is applied in the opposite direction to train the source~$\rightarrow$~target model.}}
\vspace{-0.3cm}
\label{fig:toy_illustration}
\end{figure*}

\section{Introduction} \label{sec:intro}
Machine Translation (MT) is a flagship of the recent successes and advances in the field of natural language processing.
Its practical applications and use as a testbed for sequence transduction algorithms have spurred renewed interest in this topic.

While recent advances have reported near human-level performance on several language pairs using neural approaches~\cite{wu2016google,msrMT18}, other studies have highlighted several open challenges~\cite{koehn17, isabelle17, senrich17}. A major challenge is the reliance of current learning algorithms on large parallel corpora. Unfortunately, the vast majority of language pairs have very little, if any, parallel data: learning algorithms need to better leverage monolingual data in order to make MT more widely applicable.

While a large body of literature has studied the use of monolingual data to boost translation performance when limited supervision is available, two recent approaches have explored the fully unsupervised setting~\cite{unsupNMTlample, unsupNMTartetxe}, relying only on monolingual corpora in each language, as in the pioneering work by~\citet{knight_acl11}. While there are subtle technical differences between these two recent works, we identify several common principles underlying their success. 

First, they carefully initialize the MT system with an inferred bilingual dictionary. Second, they leverage strong language models, via training the sequence-to-sequence system~\cite{sutskever2014sequence, attentionNMT} as a denoising autoencoder~\cite{vincent2008extracting}. 
Third, they turn the unsupervised problem into a supervised one by automatic generation of sentence pairs via \textit{back-translation}~\cite{sennrich2015improving}, i.e., the source-to-target model is 
applied to source sentences to generate inputs for training the target-to-source model, and vice versa.
Finally, they constrain the latent representations produced by the encoder to be shared across the two languages.
Empirically, these methods achieve remarkable results considering the fully unsupervised setting; for instance,  about 15 BLEU points on the WMT'14 English-French benchmark.

The first contribution of this paper is a model that combines these two previous neural approaches, simplifying the architecture and loss function while still following the above mentioned principles. The resulting model outperforms previous approaches and is both easier to train and tune. Then, we apply the same ideas and methodology to a traditional phrase-based statistical machine translation (PBSMT) system~\cite{pbsmt}. PBSMT models are well-known to outperform neural models when labeled data is scarce because they merely count occurrences, whereas neural models typically fit hundred of millions of parameters to learn distributed representations, which may generalize better when data is abundant but is prone to overfit when data is scarce. Our PBSMT model is simple, easy to interpret, fast to train and often achieves similar or better results than its NMT counterpart. We report gains of up to +10 BLEU points on widely used benchmarks when using our NMT model, and up to +12 points with our PBSMT model. 
Furthermore, we apply these methods to distant and low-resource languages, like English-Russian, English-Romanian and English-Urdu, and report competitive performance against both semi-supervised and supervised baselines.

\section{Principles of Unsupervised MT} 
\label{sec:principles}

Learning to translate with only monolingual data is an ill-posed task, since there are potentially many ways to associate target with source sentences.
Nevertheless, there has been exciting progress towards this goal in recent years, as discussed in the related work of Section~\ref{sec:related}.
In this section, we abstract away from the specific assumptions made by each prior work and instead focus on identifying the common principles underlying unsupervised MT.

We claim that unsupervised MT can be accomplished by leveraging the three components illustrated in Figure~\ref{fig:toy_illustration}:
(i) suitable initialization of the translation models, (ii) language modeling and (iii) iterative back-translation. In the following, we describe each of these components and later discuss how they can be better instantiated in both a neural model and phrase-based model.

\paragraph{Initialization: }
Given the ill-posed nature of the task, model initialization expresses a natural prior over the space of solutions we expect to reach, jump-starting 
the process by leveraging approximate translations of words, short phrases or even sub-word units~\cite{sennrich2015neural}. For instance, \citet{klementiev12EACL} used a provided bilingual dictionary, while \citet{unsupNMTlample} and \citet{unsupNMTartetxe} used dictionaries inferred in an unsupervised way~\cite{wordalign17, artetxeWord17}. The motivating intuition is that
while such initial ``word-by-word'' translation may be poor if languages or corpora are not closely related, it still preserves some of the original semantics.

\paragraph{Language Modeling: } Given large amounts of monolingual data, we can train language models on both source and target languages. These models express a data-driven prior about how sentences should read in each language, and they improve the quality of the translation models by performing local substitutions and word reorderings.

\paragraph{Iterative Back-translation: } The third principle is back-translation~\cite{sennrich2015improving}, which is perhaps the most effective way to leverage monolingual data in a semi-supervised setting. Its application in the unsupervised setting is to couple the source-to-target translation system with a backward model translating from the target to source language. The goal of this model is to generate a source sentence for each target sentence in the monolingual corpus. This turns the daunting unsupervised problem  into a supervised learning task, albeit with noisy source sentences. As the original model gets better at translating, we use the current model to improve the back-translation model, resulting in a coupled system trained with an iterative algorithm~\cite{he2016dual}.

\begin{algorithm}[tb]
{\footnotesize
    \SetAlgoLined
    \textbf{Language models: }Learn language models $P_s$ and $P_t$ over source and target languages\;
    \textbf{Initial translation models: }Leveraging $P_s$ and $P_t$, learn two initial translation models, one in each direction: $P^{(0)}_{s\rightarrow t}$ and $P^{(0)}_{t\rightarrow s}$\;
    \For{k=1 \textbf{to} N}{
        \textbf{Back-translation: }Generate source and target sentences using the current translation models, $P^{(k-1)}_{t\rightarrow s}$ and $P^{(k-1)}_{s\rightarrow t}$, factoring in language models, $P_s$ and $P_t$\;
        Train new translation models $P^{(k)}_{s\rightarrow t}$ and $P^{(k)}_{t\rightarrow s}$ using the generated sentences and  leveraging $P_s$ and $P_t$\;
    }}
    \caption{Unsupervised MT \label{algomain}}
\end{algorithm}

\section{Unsupervised MT systems}
\label{sec:models}

Equipped with the three principles detailed in Section~\ref{sec:principles}, we now discuss how to effectively combine them in the context of a NMT model (Section~\ref{sec:nmt}) and PBSMT model (Section~\ref{sec:pbsmt}).

In the reminder of the paper, we denote the space of source and target sentences by $\mathcal{S}$ and $\mathcal{T}$, respectively, and the language models trained on source and target monolingual datasets by $P_s$ and $P_t$, respectively. Finally, we denote by $P_{s\rightarrow t}$ and $P_{t\rightarrow s}$ the translation models from source to target and vice versa. 
An overview of our approach is given in Algorithm~\ref{algomain}.

\subsection{Unsupervised NMT} \label{sec:nmt}
We now introduce a new unsupervised NMT method, which is derived from earlier work by~\citet{unsupNMTartetxe} and~\citet{unsupNMTlample}. We first discuss how the previously mentioned three key principles are instantiated in our work, and then introduce an additional property of the system, the sharing of internal representations across languages, which is specific and critical to NMT. From now on, we assume that a NMT model consists of an encoder and a decoder. Section~\ref{sec:experiments} gives the specific details of this architecture.

\paragraph{Initialization: }
While prior work relied on  bilingual dictionaries, here we propose a more effective and simpler approach which is particularly suitable for related languages.\footnote{For unrelated languages, we need to infer a dictionary to properly initialize the embeddings~\cite{wordalign17}.} First, instead of considering words, we consider byte-pair encodings (BPE)~\cite{sennrich2015neural}, which have two major advantages: they reduce the vocabulary size and they eliminate the presence of unknown words in the output translation. Second, instead of learning an explicit mapping between BPEs in the source and target languages, we define BPE tokens by \textit{jointly} processing both monolingual corpora. If languages are related, they will naturally share a good fraction of BPE tokens, which eliminates the need to infer a bilingual dictionary. In practice, we i) join the monolingual corpora, ii) apply BPE tokenization on the resulting corpus, and iii) learn token embeddings~\citep{mikolov2013distributed}  on the same corpus, which are then used to initialize the lookup tables in the encoder and decoder. 

\paragraph{Language Modeling: } In NMT, language modeling is accomplished via denoising autoencoding, by minimizing:
\begin{eqnarray}
    \mathcal{L}^{lm} & = & \mathbb{E}_{x \sim \mathcal{S}} [-\log P_{s\rightarrow s}(x|C(x))] + \nonumber \\ 
    & & \mathbb{E}_{y \sim \mathcal{T}}[-\log P_{t\rightarrow t}(y|C(y))] \label{eq:lmNMT}
\end{eqnarray}
where $C$ is a noise model with some words dropped and swapped as in~\citet{unsupNMTlample}. $P_{s\rightarrow s}$ and $P_{t\rightarrow t}$ are  the composition of encoder and decoder both operating on the source and target sides, respectively. 

\paragraph{Back-translation: }
Let us denote by $u^*(y)$ the sentence in the source language inferred from $y \in \mathcal{T}$ such that $ u^*(y) = \arg\max P_{t\rightarrow s}(u|y)$. Similarly, let us denote by $v^*(x)$ the sentence in the target language inferred from $x \in \mathcal{S}$ such that $v^*(x) = \arg\max P_{s\rightarrow t}(v|x)$. The pairs $(u^*(y) ,y)$ and $(x,v^*(x)))$ constitute automatically-generated parallel sentences which, following the back-translation principle, can be used to train the two MT models by minimizing the following loss:
\begin{eqnarray}
    \mathcal{L}^{back} & = &
    \mathbb{E}_{y \sim \mathcal{T}} [-\log P_{s\rightarrow t}(y|u^*(y)) ] + \nonumber \\
    & & \mathbb{E}_{x \sim \mathcal{S}} [-\log P_{t\rightarrow s}(x|v^*(x))].  \label{eq:btNMT}
\end{eqnarray}
Note that when minimizing this objective function we do not back-prop through the reverse model which generated the data, both for the sake of simplicity and because we did not observe improvements when doing so. 
The objective function minimized at every iteration of stochastic gradient descent, is simply the sum of $\mathcal{L}^{lm}$ in Eq.~\ref{eq:lmNMT} and $\mathcal{L}^{back}$ in Eq.~\ref{eq:btNMT}. To prevent the model from cheating by using different subspaces for the language modeling and translation tasks, we add an additional constraint which we discuss next.

\paragraph{Sharing Latent Representations: } A shared encoder representation acts like an interlingua, which is translated in the decoder target language regardless of the input source language. This ensures that the benefits of language modeling, implemented via the denoising autoencoder objective, nicely transfer to translation from noisy sources and eventually help the NMT model to translate more fluently. In order to share the encoder representations, we share all encoder parameters (including the embedding matrices since we perform joint tokenization) across the two languages to ensure that the latent representation of the source sentence is robust to the source language. Similarly, we share the decoder parameters across the two languages. While sharing the encoder is critical to get the model to work, sharing the decoder simply induces useful regularization. Unlike prior work~\cite{gmt17}, the first token of the decoder specifies the language the module is operating with, while the encoder does not have any language identifier.

\subsection{Unsupervised PBSMT} \label{sec:pbsmt}
In this section, we discuss how to perform unsupervised machine translation using a Phrase-Based Statistical Machine Translation (PBSMT) system~\cite{pbsmt} as the underlying backbone model. Note that PBSMT models are known to perform well on low-resource language pairs, and are therefore a potentially good alternative to neural models in the unsupervised setting.

When translating from $x$ to $y$, a PBSMT system scores $y$ according to: $\arg\max_y P(y | x) = \arg\max_y P(x | y) P(y)$, where  $P(x|y)$ is derived from so called ``phrase tables'', and $P(y)$ is the score assigned by a language model.

Given a dataset of bitexts, PBSMT first infers an alignment between source and target phrases. It  then populates phrase tables, whose entries store the probability that a certain n-gram in the source/target language is mapped to another n-gram in the target/source language.

In the unsupervised setting, we can easily train a language model on monolingual data, but it is less clear how to populate the phrase tables, which are a necessary component for good translation. Fortunately, similar to the neural case, the principles of Section~\ref{sec:principles} are effective to solve this problem.

\paragraph{Initialization: }
We populate the initial phrase tables (from source to target and from target to source) using an inferred bilingual dictionary built from monolingual corpora using the method proposed by~\citet{wordalign17}. In the following, we will refer to phrases as single words, but the very same arguments trivially apply to longer n-grams. Phrase tables are populated with the scores of the translation of a source word to:
\begin{eqnarray}
p(t_j | s_i) = \frac{ e^{\frac{1}{T} \cos(e(t_j), W e(s_i)) }  }{ \sum_k e^{\frac{1}{T} \cos(e(t_k), W e(s_i)) }    }, \label{eq:uniscores}
\end{eqnarray}
where $t_j$ is the $j$-th word in the target vocabulary and $s_i$ is the $i$-th word in the source vocabulary, $T$ is a hyper-parameter used to tune the peakiness of the distribution\footnote{We set $T=30$ in all our experiments, following the setting of~\citet{inverted-sm}.}, $W$ is the rotation matrix mapping the source embeddings into the target embeddings~\cite{wordalign17}, and $e(x)$ is the embedding of $x$. 

\paragraph{Language Modeling: }
Both in the source and target domains we learn smoothed n-gram language models using KenLM~\cite{heafield2011kenlm}, although neural models could also be considered. These remain fixed throughout training iterations.

\paragraph{Iterative Back-Translation: }
To jump-start the iterative process, we use the unsupervised phrase tables and the language model on the target side to construct a seed PBSMT. We then use this model to translate the source monolingual corpus into the target language (back-translation step). Once the data has been generated, we train a PBSMT in supervised mode to map the generated data back to the original source sentences. Next, we perform both generation and training process but in the reverse direction. We repeat these steps as many times as desired (see Algorithm~\ref{alg:unsupPBSMT} in Section~\ref{sec:supplemental}).

Intuitively, many entries in the phrase tables are not correct because the input to the PBSMT at any given point during training is noisy. Despite that, the language model may be able to fix some of these mistakes at generation time. As long as that happens, the translation improves, and with that also the phrase tables at the next round. There will be more entries that correspond to correct phrases, which makes the PBSMT model stronger because it has bigger tables and it enables phrase swaps over longer spans.

\section{Experiments} \label{sec:experiments}
We first describe the datasets and experimental protocol we used. Then, we compare the two proposed unsupervised approaches to earlier attempts, to semi-supervised methods and to the very same models but trained with varying amounts of labeled data. We conclude with an ablation study to understand the relative importance of the three principles introduced in Section~\ref{sec:principles}.

\subsection{Datasets and Methodology} \label{sec:datasets}
We consider five language pairs: English-French, English-German, English-Romanian, English-Russian and English-Urdu. The first two pairs are used to compare to recent work on unsupervised MT~\cite{unsupNMTartetxe, unsupNMTlample}. The last three pairs are instead used to test our PBSMT unsupervised method on truly low-resource pairs~\cite{gu18} or unrelated languages that do not even share the same alphabet.

For English, French, German and Russian, we use all available sentences from the WMT monolingual News Crawl datasets from years 2007 through 2017. For Romanian, the News Crawl dataset is only composed of 2.2 million sentences, so we augment it with the monolingual data from WMT'16, resulting in 2.9 million sentences. In Urdu, we use the dataset of~\citet{jawaid2014tagged}, composed of about 5.5 million monolingual sentences. We report results on \textit{newstest} 2014 for $en-fr$, and \textit{newstest} 2016 for $en-de$, $en-ro$ and $en-ru$. For Urdu, we use the LDC2010T21 and LDC2010T23 corpora each with about $1800$ sentences as validation and test sets, respectively.

We use Moses scripts~\citep{moses} for tokenization. NMT is trained with 60,000 BPE codes. PBSMT is trained with true-casing, and by removing diacritics from Romanian on the source side to deal with their inconsistent use across the monolingual dataset~\cite{sennrich2016edinburgh}.

\insertphrasetable
\subsection{Initialization} \label{sec:exper_init}
Both the NMT and PBSMT approaches require either cross-lingual BPE embeddings (to initialize the shared lookup tables) or n-gram embeddings (to initialize the phrase table). We generate embeddings using fastText~\cite{bojanowski2016enriching} with an embedding dimension of 512, a context window of size 5 and 10 negative samples. For NMT, fastText is applied on the concatenation of source and target corpora, which results in cross-lingual BPE embeddings.

For PBSMT, we generate n-gram embeddings on the source and target corpora independently, and align them using the MUSE library~\cite{wordalign17}.
Since learning unique embeddings of every possible phrase would be intractable, we consider the most frequent 300,000 source phrases, and align each of them to its $200$ nearest neighbors in the target space, resulting in a phrase table of 60 million phrase pairs which we score using the formula in Eq.~\ref{eq:uniscores}. 

In practice, we observe a small but significant difference of about 1 BLEU point using a phrase table of bigrams compared to a phrase table of unigrams, but did not observe any improvement using longer phrases. Table~\ref{tab:phrase_table} shows an extract of a French-English unsupervised phrase table, where we can see that unigrams are correctly aligned to bigrams, and vice versa.

\subsection{Training}

The next subsections provide details about the architecture and training procedure of our models.

\subsubsection{NMT}
In this study, we use NMT models built upon LSTM~\cite{hochreiter1997long} and Transformer~\cite{vaswani2017attention} cells.
For the LSTM model we use the same architecture as in~\citet{unsupNMTlample}. For the Transformer, we use 4 layers both in the encoder and in the decoder. Following \citet{press2016using}, we share all lookup tables between the encoder and the decoder, and between the source and the target languages. The dimensionality of the embeddings and of the hidden layers is set to 512. We used the Adam optimizer \cite{kingma2014adam} with a learning rate of $10^{-4}$, $\beta_1 = 0.5$, and a batch size of $32$. At decoding time, we generate greedily.

\subsubsection{PBSMT}
The PBSMT uses Moses' default smoothed n-gram language model with phrase reordering disabled during the very first generation.  PBSMT is trained in a iterative manner using Algorithm~\ref{alg:unsupPBSMT}. At each iteration, we translate 5 million sentences randomly sampled from the monolingual dataset in the source language. Except for initialization, we use phrase tables with phrases up to length $4$. 

\begin{figure}[t]
\includegraphics[width=0.9\linewidth]{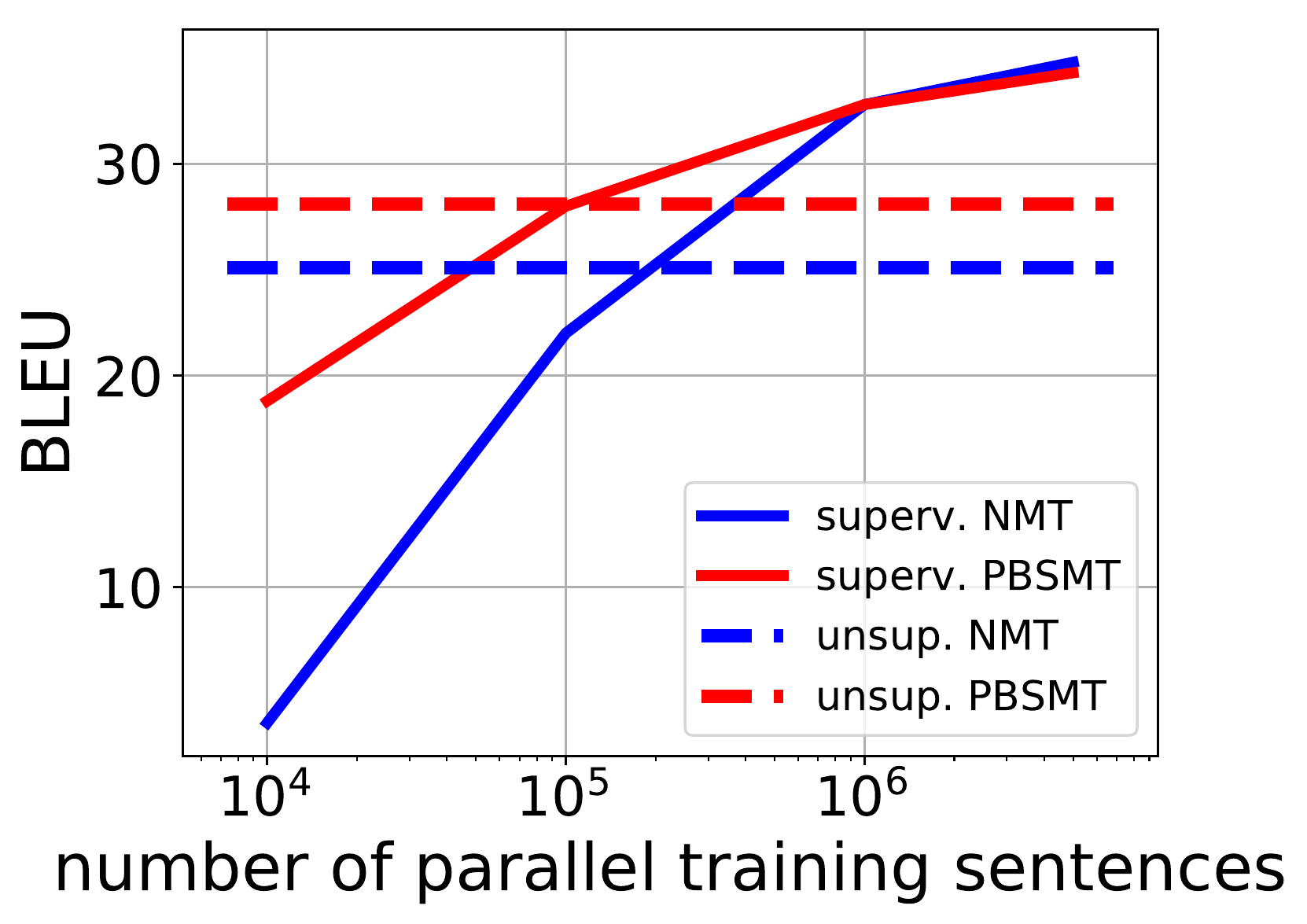}
\vspace{-0.4cm}
   \caption{\small{Comparison between supervised and unsupervised approaches on WMT'14 En-Fr, as we vary the number of parallel sentences for the supervised methods. }}
    \label{fig:supVSunsup}
\vspace{-0.5cm}
\end{figure}

\subsection{Model selection}
\label{ms}
Moses' implementation of PBSMT has 15 hyper-parameters, such as relative weighting of each scoring function, word penalty, etc. 
In this work, we consider two methods to set these hyper-parameters. We either set them to their default values in the toolbox, or we set them using a small validation set of parallel sentences. 
It turns out that with only 100 labeled sentences in the validation set, PBSMT would overfit to the validation set. For instance, on $en \rightarrow fr$, PBSMT tuned on 100 parallel sentences obtains a BLEU score of 26.42 on \textit{newstest} 2014, compared to 27.09 with default hyper-parameters, and 28.02 when tuned on the 3000 parallel sentences of \textit{newstest} 2013. Therefore, unless otherwise specified, all PBSMT models considered in the paper use default hyper-parameter values, and do not use any parallel resource whatsoever.

For the NMT, we also consider two model selection procedures: an \textit{unsupervised criterion} based on the BLEU score of a ``round-trip'' translation (source $\rightarrow$ target $\rightarrow$ source and target $\rightarrow$ source $\rightarrow$ target) as in~\citet{unsupNMTlample}, and 
cross-validation using a small validation set with 100 parallel sentences. In our experiments, we found the unsupervised criterion to be highly correlated with the test metric when using the Transformer model, but not always for the LSTM. Therefore, unless otherwise specified, we select the best LSTM models using a small validation set of 100 parallel sentences, and the best Transformer models with the unsupervised criterion.

\subsection{Results} \label{sec:results}

\insertmodelcomparisontable

\insertallresults
The results reported in Table~\ref{tab_comparison} show that our unsupervised NMT and PBSMT systems largely outperform previous unsupervised baselines. We report large gains on all language pairs and directions. For instance, on the $en \rightarrow fr$ task, our unsupervised PBSMT obtains a BLEU score of 28.1, outperforming the previous best result by more than 11 BLEU points. Even on a more complex task like $en \rightarrow de$, both PBSMT and NMT surpass the baseline score by more than 10 BLEU points. Even before iterative back-translation, the PBSMT model significantly outperforms previous approaches, and can be trained in a few minutes.

Table~\ref{tab:allresults} illustrates the quality of the PBSMT model during the iterative training process. For instance, 
the $fr \rightarrow en$ model obtains a BLEU score of 17.5 at iteration~0 -- i.e. after the unsupervised phrase table construction -- while it achieves a score of 27.2 at iteration~4. This highlights the importance of multiple back-translation iterations. The last rows of Table~\ref{tab:allresults} also show that we get additional gains by further tuning the NMT model on the data generated by PBSMT (PBSMT + NMT). We simply add the data generated by the unsupervised PBSMT system to the back-translated data produced by the NMT model. By combining PBSMT and NMT, we achieve BLEU scores of 20.2 and 25.2 on the challenging $en \rightarrow de$ and $de \rightarrow en$ translation tasks. While we also tried bootstraping the PBSMT model with back-translated data generated by a NMT model (NMT + PBSMT), this did not improve over PBSMT alone.

Next, we compare to fully supervised models. Figure~\ref{fig:supVSunsup} shows the performance of the same architectures trained in a fully supervised way using parallel training sets of varying size. The unsupervised PBSMT model achieves the same performance as its supervised counterpart trained on more than 100,000 parallel sentences.

This is confirmed on low-resource languages.
In particular, on $ro \rightarrow en$, our unsupervised PBSMT model obtains a BLEU score of 23.9, outperforming \citet{gu18}'s method by 1 point, despite its use of 6,000 parallel sentences, a seed dictionary, and a multi-NMT system combining parallel resources from 5 different languages. 

On Russian, our unsupervised PBSMT model obtains a BLEU score of 16.6 on $ru \rightarrow en$, showing that this approach works reasonably well on distant languages. 
Finally we train on $ur \rightarrow en$, which is both low resource and distant. In a supervised mode, PBSMT using the noisy and out-of-domain 800,000 parallel sentences from~\citet{tiedemann2012parallel} achieves a BLEU score of 9.8. Instead, our unsupervised PBSMT system achieves 12.3 BLEU using only a validation set of 1800 sentences to tune Moses hyper-parameters.

\subsection{Ablation Study} \label{sec:ablation}
\begin{figure*}[!t]
\minipage{0.325\textwidth}
  \includegraphics[width=\linewidth]{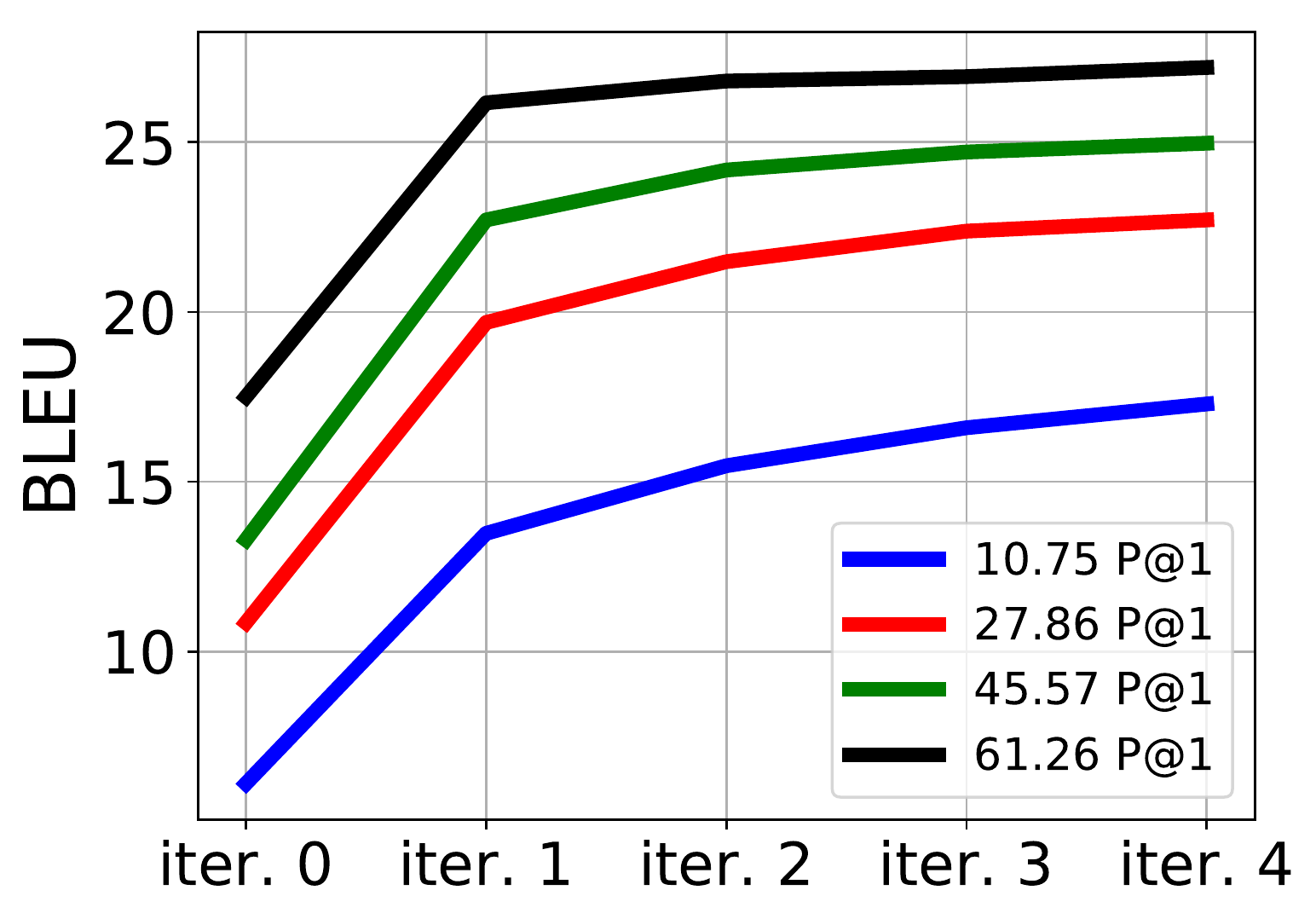}
\endminipage\hfill
\minipage{0.34\textwidth}
  \includegraphics[width=\linewidth]{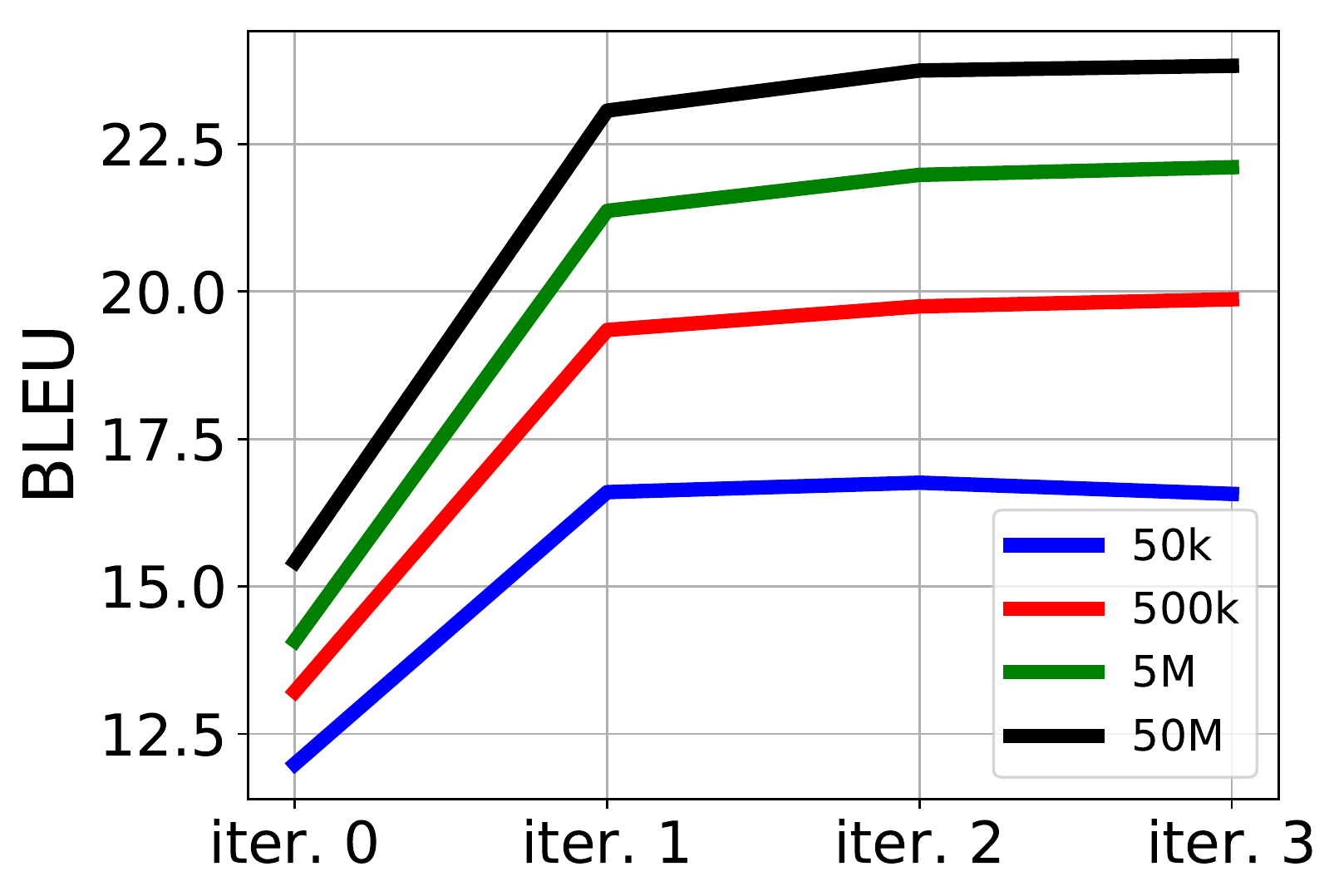}
\endminipage\hfill
\minipage{0.325\textwidth}
  \includegraphics[width=\linewidth]{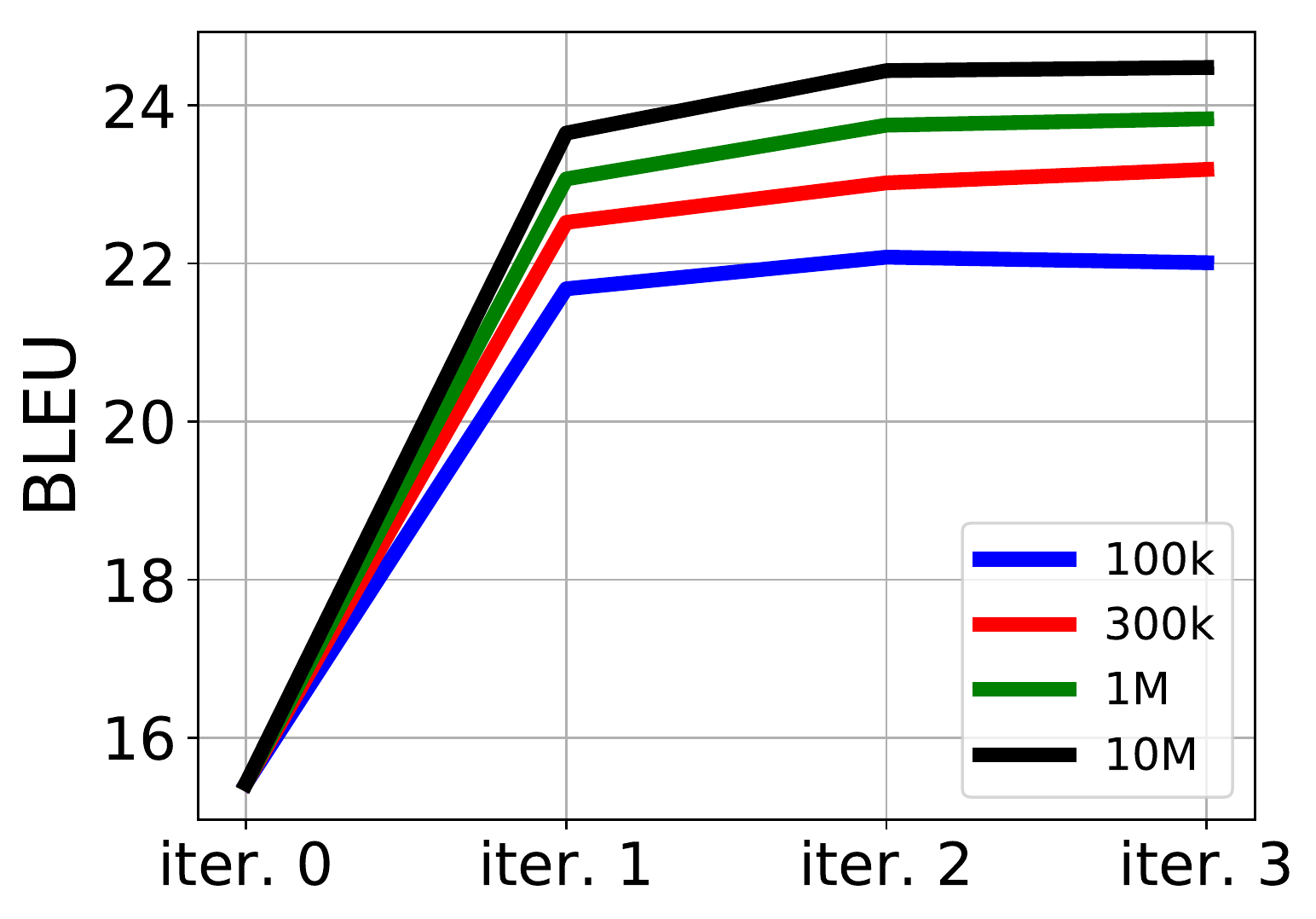}
\endminipage
\caption{\small{Results with PBSMT on the $fr \rightarrow en$ pair at different iterations. We vary: Left) the quality of the initial alignment between the source and target embeddings (measured in P@1 on the word translation task), Middle) the number of sentences used to train the language models, Right) the number of sentences used for back-translation.}}
\label{fig:ablationPlots}
\vspace{-0.25cm}
\end{figure*}

In Figure~\ref{fig:ablationPlots} we report results from an ablation study, to better understand the importance of the three principles when training PBSMT. This study shows that more iterations only partially compensate for lower quality phrase table initialization (Left), language models trained over less data (Middle) or less monolingual data (Right). Moreover, the influence of the quality of the language model becomes more prominent as we iterate.
These findings suggests that better initialization methods and more powerful language models may further improve our results.

We perform a similar ablation study for the NMT system (see Appendix). We find that back-translation and auto-encoding are critical components, without which the system fails to learn. We also find that initialization of embeddings is very important, and we gain 7 BLEU points compared to prior work~\cite{unsupNMTartetxe,unsupNMTlample} by learning BPE embeddings over the concatenated monolingual corpora.

\section{Related Work} \label{sec:related}
A large body of literature has studied using monolingual data to boost translation performance when limited supervision is available. This limited supervision is typically provided as a small set of parallel sentences~\cite{sennrich2015improving, gulcehre2015using, he2016dual, gu18, wang18}; large sets of parallel sentences in related languages \cite{firat16, gmt17, chen17acl, zheng17}; cross-lingual dictionaries \cite{klementiev12EACL, irvine14, irvine15}; or comparable corpora \cite{marcu04, irvine13}.

Learning to translate \textit{without} any form of supervision has also attracted interest, but is challenging. In their seminal work, \citet{knight_acl11} leverage linguistic prior knowledge to reframe the unsupervised MT task as deciphering and demonstrate the feasibility on short sentences with limited vocabulary. Earlier work by~\citet{carbonell06} also aimed at unsupervised MT, but leveraged a bilingual dictionary to seed the translation. Both works rely on a language model on the target side to correct for translation fluency.

Subsequent work~\cite{klementiev12EACL, irvine14, irvine15} relied on bilingual dictionaries, small parallel corpora of several thousand sentences, and linguistically motivated features to prune the search space.
\citet{irvine14} use monolingual data to expand phrase tables learned in a supervised setting. In our work we also expand phrase tables, but we initialize them with an inferred bilingual n-gram dictionary, following work from the connectionist community aimed at improving PBSMT with neural models~\cite{schwenk12, Kalchbrenner13, cho-al-emnlp14}.

In recent years back-translation has become a popular method of augmenting training sets with monolingual data on the target side~\cite{sennrich2015improving},
and has been integrated in the ``dual learning'' framework of~\citet{he2016dual} and subsequent extensions~\cite{wang18}.
Our approach is similar to the dual learning framework, except that in their model gradients are backpropagated through the reverse model and they pretrain using a relatively large amount of labeled data, whereas our approach is fully unsupervised.

Finally, our work can be seen as an extension of recent studies~\cite{unsupNMTlample,unsupNMTartetxe,unsupNMTyang} on {\it fully unsupervised} MT with two major contributions. First, we propose a much simpler and more effective initialization method for related languages. Second, we abstract away three principles of unsupervised MT and apply them to a PBSMT, which even outperforms the original NMT. Moreover, our results show that the combination of PBSMT and NMT achieves even better performance.

\section{Conclusions and Future Work} \label{sec:conclusions}

In this work, we identify three principles underlying recent successes in fully unsupervised MT and show how to apply these principles to PBSMT and NMT systems.
We find that PBSMT systems often outperform NMT systems in the fully unsupervised setting, and that by combining these systems we can greatly outperform previous approaches from the literature.
We apply our approach to several popular benchmark language pairs, obtaining state of the art results, and to several low-resource and under-explored language pairs.

It's an open question whether there are more effective instantiations of these principles or other principles altogether, and under what conditions our iterative process is guaranteed to converge. Future work may also extend to the semi-supervised setting.

\bibliography{emnlp2018}
\bibliographystyle{acl_natbib_nourl}

\clearpage
\appendix

\section{Supplemental Material}\label{sec:supplemental}
In this Appendix, we report the detailed algorithm for unsupervised PBSMT, a detailed ablation study using NMT and conclude with some example translations.

\begin{algorithm}[h!]
\SetAlgoLined
Learn bilingual dictionary using~~\citet{wordalign17}\;
Populate phrase tables using Eq.~\ref{eq:uniscores} and learn a language model to build $P^{(0)}_{s\rightarrow t}$\;
Use $P^{(0)}_{s\rightarrow t}$ to translate the source monolingual dataset, yielding $\mathcal{D}^{(0)}_{\mbox{t}}$\;
\For{i=1 \textbf{to} N}{
    Train model $P^{(i)}_{t\rightarrow s}$ using $\mathcal{D}^{(i-1)}_{\mbox{t}}$\;
    Use $P^{(i)}_{t\rightarrow s}$ to translate the target monolingual dataset, yielding $\mathcal{D}^{(i)}_{\mbox{s}}$\;
    Train model $P^{(i)}_{s\rightarrow t}$  using $\mathcal{D}^{(i)}_{\mbox{s}}$\;
    Use $P^{(i)}_{s\rightarrow t}$ to translate the source monolingual dataset, yielding $\mathcal{D}^{(i)}_{\mbox{t}}$\;
}
\caption{Unsupervised PBSMT}
\label{alg:unsupPBSMT}
\end{algorithm}

\vspace{-0.05cm}
\subsection{NMT Ablation study}

In Table~\ref{tab:abl} we report results from an ablation study we performed for NMT using the Transformer architecture.
All results are for the $en \rightarrow fr$ task.

First, we analyze the effect of different initialization methods for the embedding matrices. If we switch from BPE tokens to words, BLEU drops by 4 points. If we used BPE but train embeddings in each language independently and then map them via MUSE~\citep{wordalign17}, BLEU drops by 3 points. Finally, compared to the word aligned procedure used by~\citet{unsupNMTlample}, based on words and MUSE, the gain is about 7 points. To stress the importance of initialization, we also report performance using random initialization of BPE embeddings. In this case, convergence is much slower and to a much lower accuracy, achieving a BLEU score of 10.5.

The table also demonstrates the critical importance of the auto-encoding and back-translation terms in the loss, and the robustness of our approach to choice of architectures.

\insertablationtable
\inserttranslationsfren
\inserttranslationsruen
\inserttranslationsdeen

\subsection{Qualitative study}
Table~\ref{tab:translations} shows example translations from the French-English \textit{newstest} 2014 dataset at different iterations of the learning algorithm for both NMT and PBSMT models. Prior to the first iteration of back-translation, using only the unsupervised phrase table, the PBSMT translations are similar to word-by-word translations and do not respect the syntax of the target language, yet still contain most of the semantics of the original sentences. As we increase the number of epochs in NMT and as we iterate for PBSMT, we observe a continuous improvement in the quality of the unsupervised translations. Interestingly, in the second example, both the PBSMT and NMT models fail to adapt to the polysemy of the French word ``langue'', which can be translated as ``tongue'' or ``language'' in English. These translations were both present in the unsupervised phrase table, but the conditional probability of ``language'' to be the correct translation of ``langue'' was very high compared to the one of ``tongue'': $P(\mathrm{language}|\mathrm{langue}) = 0.92$, while $P(\mathrm{tongue}|\mathrm{langue}) = 0.0005$.
As a comparison, the phrase table of a Moses model trained in a supervised way contains $P(\mathrm{language}|\mathrm{langue}) = 0.633, P(\mathrm{tongue}|\mathrm{langue}) = 0.0076$, giving a higher probability for ``langue'' to be properly translated.
This underlines the importance of the initial unsupervised phrase alignment procedure, as it was shown in Figure~\ref{fig:toy_illustration}.

Finally, in Table~\ref{tab:translationsRu} we report a random subset of test sentences translated from Russian to English, showing that the model mostly retains the semantics, while making some mistakes in the grammar as well as in the choice of words and entities. In Table~\ref{tab:translations_deen}, we show examples of translations from German to English with PBSMT, NMT, and PBSMT+NMT to show how the combination of these models performs better than them individually.

\clearpage

\end{document}